# Half-empty or half-full? A Hybrid Approach to Predict Recycling Behavior of Consumers to Increase Reverse Vending Machine Uptime

Jannis Walk[1](✉), Robin Hirt[1], Niklas Kühl[1] and Erik R. Hersløv[2]

[1] Karlsruhe Service Research Institute, Karlsruhe Institute of Technology,
Kaiserstraße 12, 76131 Karlsruhe, Germany
{walk, robin.hirt, kuehl}@kit.edu
[2] TOMRA Systems ASA, Drengsrudhagen 2, 1385 Asker, Norway
{erik.reinhardt.herslov}@tomra.com

**Abstract.** Reverse Vending Machines (RVMs) are a proven instrument for facilitating closed-loop plastic packaging recycling. A good customer experience at the RVM is crucial for a further proliferation of this technology. Bin full events are the major reason for Reverse Vending Machine (RVM) downtime at the world leader in the RVM market. The paper at hand develops and evaluates an approach based on machine learning and statistical approximation to foresee bin full events and, thus increase uptime of RVMs. Our approach relies on forecasting the hourly time series of returned beverage containers at a given RVM. We contribute by developing and evaluating an approach for hourly forecasts in a retail setting – this combination of application domain and forecast granularity is novel. A trace-driven simulation confirms that the forecasting-based approach leads to less downtime and costs than naïve emptying strategies.

## 1 Introduction

According to the Ellen MacArthur Foundation [1], 78 million tons of plastic packaging were produced in 2013. Only 2% of this material went to closed-loop recycling, i.e. it was recycled into same or similar-quality applications. It is projected that by 2050 the oceans will contain more plastic than fish, by weight. This is a serious threat regarding food safety, nature and the world economy.

Reverse Vending Machines (RVMs) are a proven instrument for facilitating a circular economy. At our cooperating firm, TOMRA Systems ASA it is internally estimated that their RVMs facilitate 27% of the global closed-loop plastic packaging recycling. A recent study from the state of New South Wales in Australia confirms the immediate effect of deposit schemes using RVMs [2]. The deposit system was introduced in December 2017 and by May 2018 drink container litter, the largest proportion of litter volume in New South Wales, has been reduced by one third.

To facilitate a good customer experience at the RVM its uptime is crucial to avoid waiting times [3]. Additionally, an unexpected downtime of an RVM might bind an employee's capacity in undetermined and inadequate situations. The research at hand is instantiated in cooperation with TOMRA Systems ASA, the world leader on the RVM market with 75% estimated market share [4]. According to an expert interview and based on internal data, full bins are currently the number one reason for downtimes of RVMs—thus, preventing customers from entering beverage containers, which can lead to a negative connotation with RVMs in particular and recycling in general. The RVM can only continue to work after the "bin full" event is handled by an employee, who needs to stop his current activity and empty the bin.

Consequently, avoiding bin full events is the biggest lever for increasing the uptime of RVMs and, thus, improving the customer experience. The goal of the research at hand is to develop and evaluate a data-driven method for reducing bin full events, thus, improving the customer experience and optimizing an employee's workflow by warning them about bin full events beforehand. Modern RVMs are communicating the number and time of returned items. In this work, historic real-world data is first used to understand the nature of consumer's recycling behavior and the occurrence of bin full events. Then, the data serves as a basis for developing and evaluating a predictive model to avoid those situations by informing an employee beforehand. In order to solve this problem, we follow a Design Science Research approach to guide the development of our artifact [5, 6]. As we aim to solve a mature problem (the prediction of bin full events) with new solutions (analyzing RVM sensor data and deriving predictive models), thus contributing with an "improvement" [7] to the body of knowledge. To guide our artifact design, we rely on justificatory knowledge [8] from the fields of time series forecasting [9] and machine learning [10–13]. To evaluate the developed artifact from three perspectives, we perform a technical experiment [14]: First, we determine the technical performance of the predictive models and compare them to benchmarks. Second, we determine the business performance of the approach to show its utility. Third, we assess the applicability of the developed artifact to data sets from other bins.

The remainder of the paper is structured as follows. First, related work regarding the forecasting task is presented in section 2. In section 3, the use case, the data itself and the data exploration are described, and the resulting tentative design is presented. Subsequently, the evaluation and results are described in section 4. Finally, we draw conclusions and directions for further research in section 5.

## 2   Related Work

The main goal of this study is to solve the problem of unexpected bin full events based on historic data that describes the number and time of returned items. Thus, this is a time series forecast problem. Usually, several bin full events per day occur on a given RVM. Hence forecasts with an hourly forecast granularity are required. Related work can be structured along the application domain, and the forecast granularity (e.g. hourly, daily or weekly periods). An excerpt of related work is shown in Table 1. Due to limited relevance of larger forecast granularities only papers addressing forecasts on hourly or daily level are included in the table.

The work at hand is very closely related to retail forecasting since RVMs are mostly placed in supermarkets. Thus, customers usually combine shopping and returning empty beverage containers. However, in the domain of retail forecasting no published work with an hourly forecasting is found. Aburto and Weber (2007) [10] and Hasin et. al (2011) [15] forecast the daily demand of various products in a supermarket. Taylor (2007) [16] forecasts the daily demand for one product. Thiesing and Vornberger (1997) [17] forecast the demand of 20 products in a supermarket on a weekly basis. Also in the financial forecasting domain no hourly forecasts are found, e.g. Kim (2003) [18] forecasts the direction of change in the daily Korean composite stock price index. Other papers like Wang et al. (2012) [19] forecast trends of indices on a monthly basis and are therefore not included in the table. Hourly forecasts are used for example in the meteorology domain: Sfetsos (2000) [20] forecasts the mean hourly wind speed series on the island of Crete. Sfetsos and Coonick (2000) [21] forecast the hourly solar radiation. However, these time series are not generated by human behavior like in the case at hand. In other application domains hourly forecasts for time series generated by human behavior exist, however our forecasting problem poses special challenges explained in the following. In the domain of bike sharing demand forecasting hourly forecasts for time series reflecting human behavior are used (compare Li et al. (2015) [22]). However, the domain is different, for example, weather has an immediate effect on bike sharing demand. This is not the case for recycling behavior. In electricity load and price forecasting hourly forecasts are already used for a long time (the interested reader finds a review of electricity price forecasting in [11]). However, the domain is different, and forecasts are in a less fine granularity regarding the spatial dimension: Fan and Chen (2006) [23] forecast the hourly electricity demand in New York City. Crespo Cuaresma et al. (2004) [24] and Kristiansen (2012) [25] forecast the hourly electricity prices for whole energy exchanges spanning several European countries. For the problem at hand we consider only single bins of single RVMs and thus the behavior of drastically fewer people. Consequently, even though both forecasts have an hourly granularity, our forecasting problem has a finer granularity.

Hitherto, hourly forecasts have not been examined in the application domain of retail or a comparable application domain. Hence, we contribute by developing and evaluating an approach for hourly forecasts in a retail setting where the time series are fine granular in the sense that they reflect the behavior of relatively few humans.

**Table 1.** Excerpt of Related Work regarding the forecasting task.

| Source | Application Domain | | | | | Forecast granularity | |
|---|---|---|---|---|---|---|---|
| | Retail | Finance | Meteorology | Bike sharing | Electricity | Hourly | Daily |
| [10] [15, 16] | X | | | | | | X |
| [18] | | X | | | | | X |
| [20, 21] | | | X | | | X | |
| [22] | | | | X | | X | |
| [23–25] | | | | | X | X | |
| This work | X | | | | | X | |

## 3   Predicting Bin Full Events by Forecasting Consumer Recycling Behavior

In this study we want to decrease the downtime of RVMs by predicting bin full events based on historic data. In the following, we first describe the use case, the data we used as a basis for our forecasting method and the data exploration. Then we present our tentative design that levers a hybrid prediction model combining machine learning and statistical approximation and is using endogenous and exogenous data.

### 3.1   Use Case, Data Description and Exploration

In practice, two data sources reflecting the fill level of a bin are available: The measurement of a sensor positioned above a bin and the time and type of all returned items. 90% fill level is the last signal from the bin level sensor before the bin full signal. Thus, relying solely on the bin level sensor would be too inefficient: The employee would need to empty the bin when there is still capacity left. Furthermore, the sensor data would be too coarse-grained as basis for a prediction model. Consequently, our envisioned system uses the bin level sensor only to know, when a fill level is at 90%. This reduces uncertainty since "the further ahead we forecast, the more uncertain we are" [9]. This 90% fill level signal will be used to trigger a forecast about the remaining 10% fill level which is based on detailed item return data. This is based on the fact that every fill level change is associated with a certain number of items going into that bin. Thus, instead of predicting the fill level directly, we predict the number of returned items. This prediction output can be mapped back to a corresponding fill level.

In addition to the data about returned items, we consider exogenous data sources, such as the weather or public holidays, as potential candidates for input features in the prediction model. Having stated which data sources are used now the data set and the results of the data exploration are presented.

The data used to build the artifact stems from one bin of one RVM in a supermarket in Norway and covers the period from the 12th of June 2014 to the 29th of May 2017. To make the data interpretable it needs to be aggregated to fixed time intervals. Thus, all item returns within these time intervals (monthly, daily and hourly) are added up. The monthly number of returned items shows an upward trend. Additionally, a monthly seasonality is present: there are more item returns in summer than in winter, which seems reasonable due to the higher temperatures in summer. These lead to a higher consumption and return of beverage containers. Also, it is observed that there are less item returns in July than in June and August, which could be explained by July being the main vacation month in Norway.

When the data is aggregated to a daily interval it can be observed that the 25% quantile of returned items on Saturdays is higher than the median of all other weekdays. The other weekdays do not show significant differences. Also, the distribution of the number of returned items per day has a long tail, i.e. on few days, an abnormal high number of returned items can be observed. In a second stage of data exploration further possible predictors for a prediction model for the daily time series are identified. The auto-correlation plot (Figure 1) shows that the number of returned items today has the strongest correlation with the number of items returned yesterday and on the same weekday 1 to 5 weeks ago (lag of 6, 12, 18, 24 and 30 business days since the supermarket considered is closed on Sundays). These correlations are larger than 0.3.

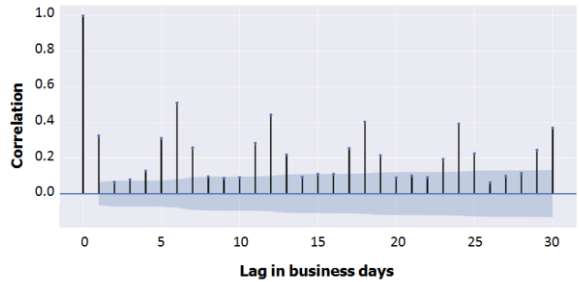

**Fig. 1.** Auto-correlation plot of daily item returns with maximum 30 business days lag

Previous research indicates that shopping behavior is influenced by the weather [26, 27]. Since recycling behavior is assumed to be very similar to shopping behavior weather variables are included in our predictive model. Another exogenous influence are holidays. It seems that this variable is also relevant for the problem at hand: The 25% quantile of days after holidays is higher than the 75% quantile of all other days. However, there are only 34 holidays in the data set. Thus, the sample size is relatively small and consequently we cannot be very confident about this statement. For days after holidays no significant difference can be found.

The hourly time series is more variable than the daily item returns time series (coefficient of variation of 125.90% vs. 43.55%). Also, it can be observed that in almost 17% of the hours no items are returned. The hourly item returns can be interpreted as observed hourly demand for returning empty beverage containers at a given store. As such it can be characterized as lumpy demand according to e.g., Gutierrez el a. (2008) [28], since it is characterized by many intervals with zero demand and periods with actual demand occurrences. It is a known problem to produce reliable forecasts for time series which are characterized by lumpy demand [29, 30]. Besides the autocorrelations for the hourly time series are all below 0.3. Overall, the higher coefficient of variation, the lumpiness and the lower autocorrelations suggest that the hourly time series is more difficult to forecast than the daily forecast. Consequently, the predictive modeling is based on a forecast of the number of returned items per day.

Based on the data exploration a data set is constructed for the predictive modeling. The number of returned items per day is the target variable of the prediction model. The first group of predictors are the lagged variables: the number of returned items 1, 6, 12, 18, 24 and 30 days ago are used. For the seasonalities date-related variables are used: the weekday as numeric variable, the month and the year. Besides two types of exogenous variables are used: boolean variables indicating a day before or after a public holiday and the precipitation intensity and the apparent maximum temperature as weather variables.

### 3.2 Tentative Design: A Hybrid Prediction Model Based on Endogenous and Exogeneous Data Sources

The tentative design is a hybrid prediction model which uses machine learning and statistical approximation and is based on different data sources. The prediction of bin full events is triggered by the bin level sensor indicating 90% bin level. Then the data set described above is used as input for a machine learning model to forecast the daily item returns. The output of this machine learning model is used to compute hourly forecasts by statistical approximation. Based on this the notification sent to the smartphone of the store employee will be generated. Figure 2 shows an overview of the envisioned tentative design. The two steps in the middle are described in the following. The first and last step are to be realized for an artifact in practice. This realization is viable: the sensors for the first step already exist. Also, there is already a smartphone app which can be used for the notification of the store employee. The envisioned productive system is illustrated in the lower part of Figure 2.

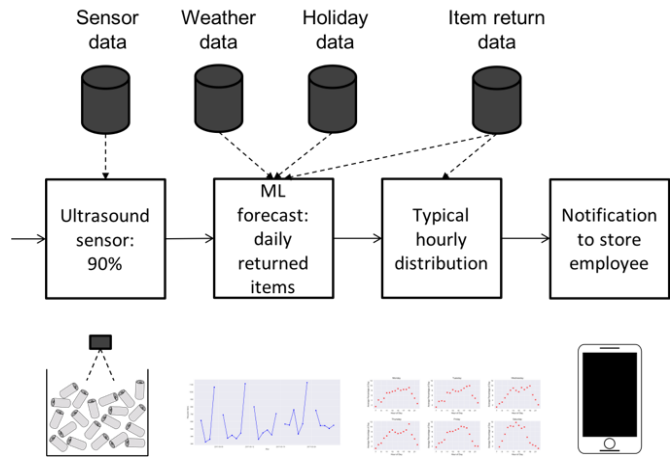

**Fig. 2.** Tentative design: A Hybrid Prediction Model to predict bin full events

In the following it is described how the forecasting model was selected. The recently most widely used techniques for time series forecasting can be categorized into time series models, like Autoregression Moving Average (compare Brockwell and Davis (2016) [31]) and machine learning models. Time Series models are specifically designed to forecast time series and usually use only information stemming directly from the time series. Machine learning models such as Neural Networks and Support Vector Machines are applicable for a wider set of tasks like classification and clustering. For time series forecasting regression algorithms are applied.

In this work, we are analyzing long series with high-frequency data, where machine learning models work better according to Bergmeir and Benítez (2012) [13]. Also, research implies that machine learning models can deal better with the complexity and non-linearity of electricity price time series than time series models [11]. According to this paper, Neural Networks, Support Vector Regression and Multiple Linear Regression have shown to produce reliable forecasts for electricity prices. Research in the closely related application domain of retail forecasting also suggests that machine

Half empty of half full?        7learning models (e.g. Neural Networks) outperform classical time series models [6]. Consequently, machine learning models, namely Support Vector Regression, Artificial Neural Networks and Multiple Linear Regression, are used for the forecasting task at hand. Additionally, a Gradient Boosting Regressor (compare Friedman (2001) [12]) is considered due to its successful application in classical research as well as data science competitions [33].

To predict bin full events, hourly forecasts are necessary. We implement and evaluate several approaches and present the most promising one. First, several approaches for direct forecasts of the number of returned items per hour are implemented and evaluated. This is a wide-spread approach in application domains like electricity forecasting [23–25]. Second, we implement and evaluate a hybrid approach based on machine learning and statistical approximation (averaging). This approach is similar to the one described by Gross and Sohl (1990) [34]. Their disaggregation based on statistical approximation is from product families to products. In the case at hand the disaggregation is in a temporal dimension: from daily forecast to hourly forecast. The second approach outperforms the direct hourly forecasts and is described in more detail in the following. As depicted in Figure 3, typical distributions for each weekday are used to convert the daily forecast into an hourly one. The upper plot shows the daily forecast. The middle plot shows the distribution which is used for mapping. It is obtained by calculating the mean of the percentage of day values for each hour of all preceding same weekdays. The daily forecast is multiplied with these percentage values to obtain the forecast for the respective hour. This resulting forecast is shown in the lower plot.

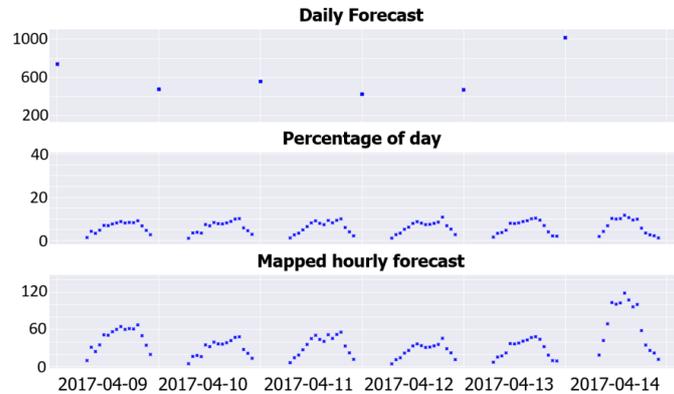

**Fig. 3.** Illustration of mapping from daily to hourly forecast

## 4  Evaluation & Results

We evaluate the presented artifact in three steps: First, we perform a technical experiment to determine the performance of the prediction model for daily and hourly returned items and compare it to forecast benchmarks. Second, we evaluate the performance of the proposed approach from a business perspective, considering the cost for RVM downtime and the cost for perished bin capacity. Third, we briefly evaluate the performance of the prediction model on other RVMs at different locations to test the degree of generalization of the approach.

### 4.1   Performance of the Prediction Model

As first evaluation measure we use the Mean Absolute Error (MAE) that is calculated by

$$\text{MAE} = \frac{\sum_{i=1}^{n}|y_i - \hat{y}_i|}{n} \qquad (1)$$

as it is easily understandable. Because it is scale-dependent, it cannot be used to compare performances across data sets. One of the most widely used scale-independent measures is the Mean Absolute Percentage Error. However, it cannot be used for the data at hand since it is not suitable if $y_i = 0$ or close to zero. Then it yields infinite or very high numbers. For data sets with this property the MAE can be divided by the mean of the actual values to obtain a scale-dependent error measure [35]. These two evaluations measures are used in the following.

In time series forecasting it is good practice to report the performance of simple forecasting methods as benchmarks in addition to the evaluation measures [9]. The following three benchmarks are reasonably applicable for the data set used in the following. A daily time series with weekly seasonality is used for illustration purposes. The naïve forecast uses the value of yesterday as forecast for today: $\hat{y}_i = y_{i-1}$. The seasonal naïve forecast is a similar approach. Its forecast for today is the value of the same day last week: $\hat{y}_i = y_{i-m}$, with $m$ = seasonality period. Another applied benchmark is the Seasonal Moving Average, it computes the forecast by averaging the value of the last $x$ same weekdays.

As data splitting strategy a cross-validation is applied since it achieves more precise error estimates for time series data than time series specific procedures [13, 36].

Table 2 shows the results of the aforementioned benchmarks and machine learning algorithms. For each forecasting algorithm which requires training 10-fold cross-validation is applied. The mean of returned items per day is 512.11.

For the Seasonal Moving Average $x = 5$ yields the best results, thus the forecast is computed by averaging the values of the same weekday 1 to 5 weeks ago. For the Support Vector Regressor all available kernels ('rbf', 'linear', 'poly' and 'sigmoid') are tested. The linear kernel performs best and is used to obtain the results in Table 2. Due to limited computing capabilities on the available hardware no grid-search for parameter tuning is conducted and therefore the standard parameters from scikit-learn [37] are used. Parameter tuning can and most probably will further enhance the performance of the machine learning algorithms.

As one can observe in Table 2 all machine learning models yield better results than the simple benchmarks. The differences between the machine learning models are relatively small. When inspecting the machine learning models, the weekday and the lagged variables 1 to 5 weeks ago are always among the most important predictors. This explains the similar performance of the Seasonal Moving Average and the machine learning models.

Table 2. Results of different benchmarks and machine learning forecasts for daily item returns.

| Method | MAE | MAE/Mean |
|---|---|---|
| Naïve forecast | 194.51 | 37.98% |
| Seasonal naïve forecast | 162.22 | 31.68% |
| Seasonal Moving Average | 136.49 | 26.65% |
| Multi-layer Perceptron Regressor | 134.85 | 26.33% |
| Support Vector Regressor | 129.66 | 25.32% |
| Multiple Linear Regression | 129.00 | 25.19% |
| Gradient Boosting Regressor | 125.08 | 24.42% |

In case the Seasonal Moving Average performs almost as good as the best performing model for more data sets it is a promising candidate for the productive use of such a system. It does not require any learning and thus facilitates the development of a productive system considerably.

The results of the hourly forecast are shown in Table 3 for a selection of algorithms. The mapping approach is the one explained in section 3.2. The mean of returned items per hour is 32.24.

Table 3. Results of different forecasting methods for hourly item returns.

| Method | MAE | MAE/Mean |
|---|---|---|
| Naïve forecast mapped to hourly forecast | 25.74 | 79.59% |
| Seasonal Moving Average mapped to hourly forecast | 24.22 | 74.89% |
| Gradient Boosting Regressor mapped to hourly forecast | 23.82 | 73.65% |

The MAE/Mean evaluation measures for hourly item returns are approximately three times worse than for the daily item returns. In the data exploration in section 3.1 the lumpiness of the hourly item returns time series is already mentioned as a factor which negatively impacts its forecastability. In addition, there is the use-case-specific challenge of bulk arrivals. A bulk arrival occurs when one customer returns a high number of empty beverage containers at once. In the data set considered the most extreme bulk arrival contains 285 items. On average just 32.24 items are returned per hour. Hence, these events which cannot be foreseen with the data at hand yield extreme spikes in the hourly time series.

The similarity of these performance values shows that the forecasting accuracy for the daily item returns has little influence on the forecasting accuracy on hourly basis. The inaccuracy introduced by mapping via past hourly distributions is the major inaccuracy. The results of the naïve forecast are reported to highlight this. The relatively large performance differences on a daily level between the Gradient Boosting Regressor and the naïve forecast do not translate into big performance differences on an hourly level.

### 4.2  Artifact Evaluation from a Business Perspective

The evaluation measures reported in the section above cannot be interpreted regarding the utility of the approach. Thus, to evaluate the artifact from a business perspective, key performance indicators reflecting the different costs of emptying strategies are selected with domain experts: the percentage of avoided bin full events and the average number of hours the notification is too early (compare Figure 4).

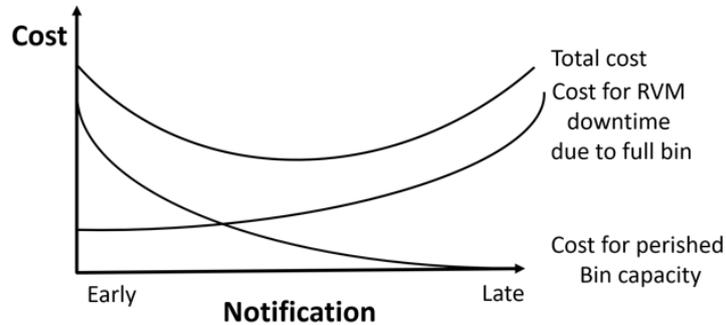

**Fig. 4.** Trade-off between different costs for bin full Notifications

A bin full event is avoided, if the employee is notified about an upcoming bin full event in advance. A simple strategy to avoid a high percentage of bin full events is to use very early notifications. Therefore, it is also evaluated how many hours before the actual bin full event the notifications are on average. This KPI can then be used to assess how much perished bin capacity and additional labor cost there is to expect for a certain combination of policy and forecast. To compare different emptying strategies regarding the aforementioned KPIs a simulation is implemented. The simulation is trace-driven, i.e. real data is used as input, compare Sargent (2013) [38].

As for the forecasting, first simple benchmarks are constructed. The benchmarks are different hour-based policies, they do not rely on any forecast. For example, the time in the notification is always two hours after the 90% bin level signal (compare the upper rows of Table 4).

**Table 4.** Simulation: Results of hour-based and forecast-based policies.

| hours between 90% bin level & Notification | Forecasting method | % avoided bin full events | Average hours too early |
|---|---|---|---|
| 2 | - | 79.90 | 3.37 |
| 0 | - | 100 | 4.28 |
| - | Naïve forecast | 77.95 | 2.38 |
| - | Seasonal moving average | 80.33 | 2.24 |
| - | Gradient Boosting Regressor | 81.17 | 2.16 |

Then, these benchmarks are compared to policies in which the time in the notification is determined based on different forecasts. For that, a data set with the actual and forecasted hourly item returns is used (since cross-validation is applied forecasts exist for each business hour of the observed period). To obtain the time in the notification, the forecasted item returns time series is accumulated until it reaches 100 items. On average the considered bin takes 131 items between 90 and 100% bin level, the 100 items are arbitrarily chosen to have a buffer. Compare the bottom rows of Table 4 for the corresponding results. For each emptying policy there were minimum 2354 simulated bin full events.

The most important observation is that the simulation shows that forecast-based policies yield better KPIs than hour-based policies. The two-hour policy and the forecast-based policies with a threshold of 100 can be compared very well. Using the Gradient Boosting Regressor and the Seasonal Moving Average leads to an improvement of both KPIs: the percentage of avoided bin full events is slightly higher while the notifications are on average more than one hour less too early. Consequently, both the cost for downtime and the cost for perished bin capacity are lower. Thus, the total cost curve is lower for forecast-based policies than for hour-based policies.

Also, it can be observed that there are only little differences between the different forecasting methods. In case of putting the artifact into practice, it needs to be assessed if the relatively small enhancement of the Gradient Boosting Regressor over the Seasonal Moving Average justifies the necessary additional effort and the potential cost for weather data from external providers.

One rather strong assumption is that the considered bin takes always exactly 131 items after a 90% bin level signal until the bin is full. This is only an average value, there will be deviations in real life. Additionally, a perfectly functioning bin level sensor is assumed in the sense that the indicated 90% bin level is accurate. It is expected that the bin level sensor works reliable, but it is not capable of detecting the current bin level precisely. As a consequence of these limitations, the results of the simulation are not to be understood as expected real life behavior. But all limitations apply both for the hour and forecast-based policies. Consequently, they provide a valid guideline regarding the difference between these two approaches. They clearly demonstrate the benefit of forecasting the number of returned items per hour instead of relying on hour-based policies.

### 4.3 Applicability to Other RVMs

So far, only one bin of one RVM is considered. To assess the generalizability the same forecasting model is trained and evaluated on the data of four more bins. The according supermarkets are chosen to have different location characteristics (city center & rural area) and to be in different countries. For the sake of brevity only the results of the hourly forecast are mentioned: the respective best forecasting approaches yield MAE/Mean values between 50,75 and 63,03%. The MAE/Mean for the time series considered so far is 73,62%. Thus, the approach works better for all additionally considered time series, which confirms the generalizability of the approach.

## 5 Conclusion & Outlook

Bin full events are the major reason for downtime of RVMs and yield waiting times for supermarket customers as well as negative customer experience—and might bind an employee's capacity in undetermined and inadequate situations. To address this problem, the paper at hand presents and evaluates a hybrid approach for predicting events of full bins by applying a Design Science Research approach. The presented hybrid approach combines machine learning and statistical approximation and uses endogenous and exogenous data sources. The obtained forecasts are not very accurate due to lumpiness and bulk arrivals negatively impacting the forecastability of the hourly item returns. However, a trace-driven simulation shows that bin emptying strategies based on the hybrid approach yield a higher utility compared to naïve preventive bin emptying strategies. Overall, we contribute by developing and evaluating an approach for hourly forecasts in a retail setting where the time series granularity reflects the behavior of relatively few humans. Such studies on forecastability are encouraged contributions both in Information Systems research in general ([39]) and in the field of reverse logistics ([40]) for product returns. Also, the paper at hand shows how machine learning and statistical approximation can be leveraged to facilitate a good customer experience and increase process efficiency. Ultimately, this enables a further proliferation of RVMs which help to increase the rate of plastic packaging being recycled in a closed-loop.

Besides the contribution, this work also has limitations. So far, the approach has only been tested for item return time series of five different bins. Evaluations on data sets from more bins are to be performed to confirm the approach. Also, the simulation assumes that the considered bin always takes exactly 131 items between 90 and 100% bin level and the sensor indicating the current bin level works faultlessly.

However, this work lays ground for future research in the field of hourly forecasts for lumpy time series. It shows that the hybrid approach combining machine learning and statistical approximation yields better results than direct hourly forecasts. This can be tested for time series with similar characteristics like supermarket sales on stock keeping unit level. Another possibility for future research is the realization of the tentative design by integrating the live bin level sensor data and the smartphone app for the notification of the supermarket employee.

Our study informs managers of supermarkets and RVM manufacturers that predicting bin full events is possible and relying on a forecasting-based approach to do so seems reasonable. This facilitates proactive bin emptying strategies instead of the currently used reactive ones. Ultimately, when implemented, the resulting artifact can facilitate customer experience with reduced waiting times—and contribute to the general experience of recycling as a positive one. Also, it can facilitate a more sustainable operation of further parts of the value chain: in some markets like Australia RVMs are placed outside supermarkets. For such RVMs the forecast can be used to optimize the schedule for the trucks picking up the empty beverage containers. Thus, the overall reverse vending logistics chain can be optimized regarding its sustainability.

Due to the limitations of the simulation, a real-world evaluation is necessary to confirm the utility of the approach. In this evaluation it should also be assessed how precisely the proposed system affects employee and customer satisfaction. For this real-world evaluation the usage of the Seasonal Moving Average for the forecast of the daily time series seems most reasonable. It considerably facilitates the development of the proposed system since no model training is necessary while the MAE/Mean measure for the hourly time series decreases only 1.24 percentage points. Later, if the system is deployed to more RVMs a more sophisticated, machine learning based method can be beneficial due to economies of scale.